\pdfoutput=1

\documentclass[11pt]{article}

\usepackage{acl}

\usepackage{times}
\usepackage{latexsym}

\usepackage[T1]{fontenc}

\usepackage[utf8]{inputenc}

\usepackage{microtype}

\usepackage[ruled,vlined]{algorithm2e}
\usepackage{graphicx}
\usepackage{booktabs}
\usepackage{amsmath}
\usepackage{bbm}
\usepackage{multirow}
\usepackage{stfloats}
\title{End-to-end Training and Decoding for \\ Pivot-based Cascaded Translation Model}

\author{
  Hao Cheng\textsuperscript{\rm{1}}\thanks{{} {} Work done during the internship at Huawei Noah's Ark Lab.}, Meng Zhang\textsuperscript{\rm{2}}, Liangyou Li\textsuperscript{\rm{2}},  {\bf  Qun Liu\textsuperscript{\rm{2}}}, {\bf Zhihua Zhang\textsuperscript{\rm{3}}} \\
  \textsuperscript{1} Academy for Advanced Interdisciplinary Studies, Peking University \\
  \textsuperscript{2} Huawei Noah's Ark Lab \\
  \textsuperscript{3} School of Mathematical Sciences, Peking University \\
  \texttt{hao.cheng@pku.edu.cn} \\
  \texttt{\{zhangmeng92, liliangyou, qun.liu\}@huawei.com} \\
  \texttt{zhzhang@math.pku.edu.cn}
}

\begin{document}
\maketitle
\begin{abstract}
Utilizing pivot language effectively can significantly improve low-resource machine translation. Usually, the two translation models, source-pivot and pivot-target, are trained individually and do not utilize the limited (source, target) parallel data. This work proposes an end-to-end training method for the cascaded translation model and configures an improved decoding algorithm. The input of the pivot-target model is modified to weighted pivot embedding based on the probability distribution output by the source-pivot model. This allows the model to be trained end-to-end. In addition, we mitigate the inconsistency between tokens and probability distributions while using beam search in pivot decoding. Experiments demonstrate that our method enhances the quality of translation.
\end{abstract}

\section{Introduction}
Neural machine translation has developed rapidly with the development of deep learning~\citep{sutskever2014sequence, bahdanau2014neural, vaswani2017attention}.
Generally, the training of these models requires a large number of parallel data. However, existing parallel data mainly focus on English, limiting the development of other language pairs. Now researchers are increasingly interested in other languages with limited resources.

Pivot-based methods effectively alleviate the problem of low resources by using a pivot language~\citep{de2006catalan, utiyama2007comparison}. The pivot language has rich parallel data with the source and target languages. Usually, the source-pivot and the pivot-target translation models are trained independently, which can not fully use a small number of parallel data between the source and target languages. \citet{ren2018triangular} jointly optimize translation models with a unified bidirectional EM algorithm. \citet{kim2019pivot} and \citet{zhang2022triangular} use the method of transfer learning, while \citet{cheng2019joint} use the method of joint optimization. In this paper, we also propose a joint optimization method.

Inspired by \citet{bahar2021tight}, we re-normalize the pivot token probability distribution of the source-pivot model output and weight the pivot word embedding as the input of the pivot-target model. In this way, we can fine-tune the two cascaded translation models end-to-end. When beam search is used in pivot decoding, the generated tokens are inconsistent with the probability distributions. We design an improved decoding algorithm to alleviate the inconsistency problem. 
We conduct extensive experiments to verify the effectiveness of our method.

\section{Methodology}
\begin{figure*}[t]
  \centering
  \includegraphics[width=0.95\linewidth]{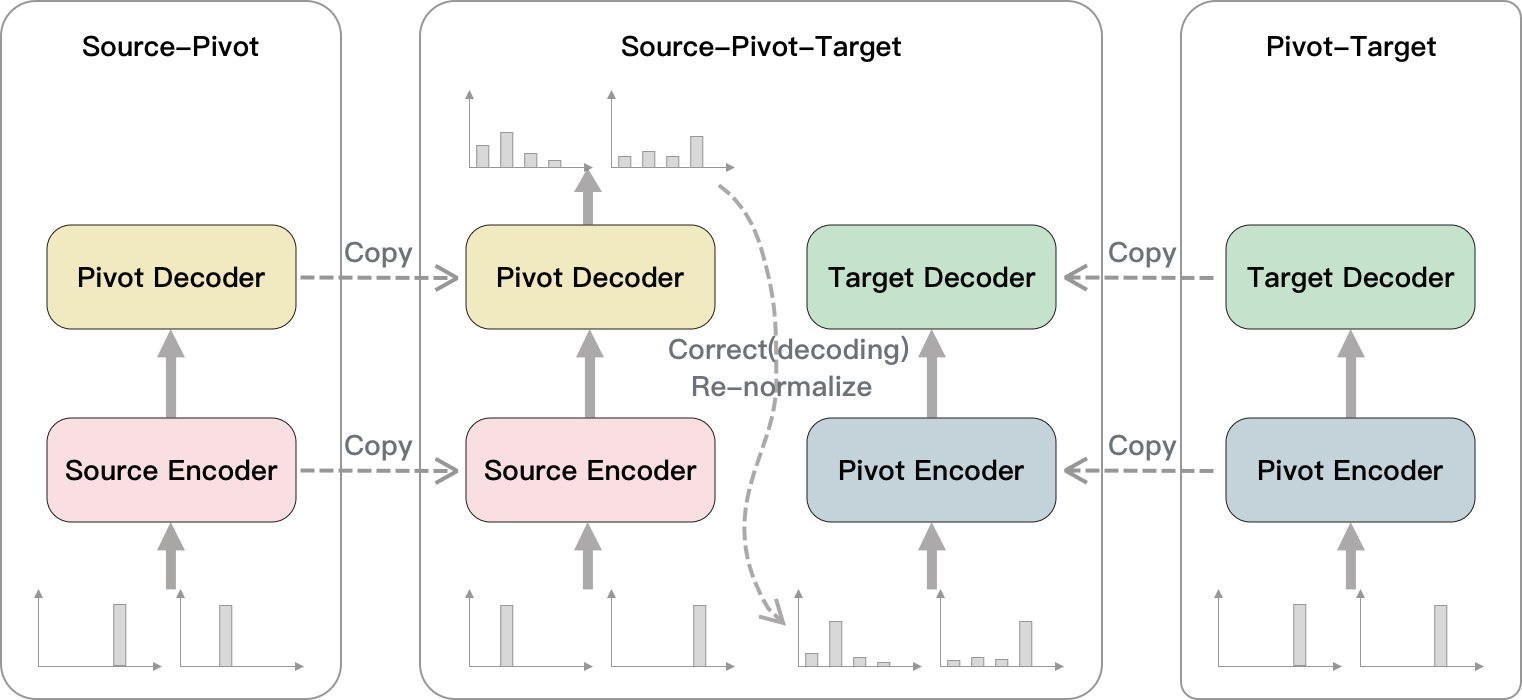}
  \caption{The illustration of our method. We use pre-trained source-pivot and pivot-target models to initialize the cascaded model source-pivot-target. The bottom/top distribution indicates that the input/output is one-hot or soft. We only correct the probability distribution in decoding.}
  \label{fig:method}
\end{figure*}

We connect two pre-trained translation models (source-pivot and pivot-target) in series to initialize our cascaded translation model source-pivot-target.

In order to train the cascaded model end-to-end, 
we collect the probability distributions of pivot tokens at each position as the additional output of source-pivot.
For the pivot-target model, we use the probability-weighted sum of embeddings in the pivot vocabulary as input instead of the embedding of a specific token, which enables backpropagation. Before weighting the embedding, we re-normalize the probability.
Figure~\ref{fig:method} shows the illustration of our method.

\subsection{Probability Re-normalization}

There is a gap between pivot encoder inputs of the pre-trained pivot-target and the source-pivot-target model. 
Therefore we try to re-normalize the probability to make it more peaked, which is closer to the one-hot vector. The re-normalized probability distribution is defined as follows:

\begin{equation}
p\left(z_{t} \mid \boldsymbol{Z_{<t}}, \boldsymbol{X} \right)=\frac{p^{\alpha}\left(z_{t} \mid \boldsymbol{Z_{<t}}, \boldsymbol{X}\right)}{\sum_{z \in\boldsymbol{V_{Z}}} p^{\alpha}\left(z \mid \boldsymbol{Z_{<t}}, \boldsymbol{X}\right)},
\end{equation}
where $\boldsymbol{X}$ denotes the source sentence, $\boldsymbol{Z_{<t}}$ denotes the pivot tokens generated before time step $t$, $\boldsymbol{V_{Z}}$ denotes the pivot vocabulary, and $\alpha$ denotes the exponent hyper-parameter. 

\subsection{Training Loss}
We use the parallel data (source, target) to train the source-pivot-target model. 
In addition, we use the pre-trained source-pivot model to translate the source data in (source, target) to pivot and get the trilingual parallel data (source, pseudo-pivot, target). We calculate losses both of pivot and target as follows:
\begin{align}
\mathcal{L}_{pivot}&=-\sum_{t} \log p\left(z_{t} \mid \boldsymbol{Z_{<t}}, \boldsymbol{X}\right),\\
\mathcal{L}_{target}&=-\sum_{t} \log p\left(y_{t} \mid \boldsymbol{Y_{<t}}, \boldsymbol{X}\right),
\end{align}
where $\boldsymbol{Y_{<t}}$ denotes the target tokens generated before time step $t$. And the final training loss is:
\begin{equation}
\label{con:loss}
\mathcal{L}= \beta\mathcal{L}_{pivot}+\gamma\mathcal{L}_{target},
\end{equation}
where $\beta$ and $\gamma$ are hyper-parameters to control the intensity of different losses.

\subsection{Training Steps}
We summarize our training steps as follows:
\begin{itemize}
\item Train the source-pivot and the pivot-target models using (source, pivot) and (pivot, target) parallel data, respectively.
\item Translate source in (source, target) using the source-pivot model to obtain the trilingual parallel data (source, pseudo-pivot, target).
\item Connect the source-pivot and the pivot-target models in series to initialize the source-pivot-target model.
\item Modify the encoder input of the pivot-target model in the source-pivot-target model.
\item Train the source-pivot-target model end-to-end on (source, pseudo-pivot, target) with $\mathcal{L}$.
\end{itemize}

\subsection{Decoding and Probability Correction}
Both greedy search and beam search can be used to decode source $\to$ pivot and pivot $\to$ target translations. 
When beam search is employed in source $\to$ pivot, both the selected token and the probability distribution are preserved in the search. 
This results in inconsistency between the output tokens and the probability distributions. 
At some positions, the token with the highest probability does not match the generated token. 
Since beam search produces the final sequence of tokens, the tokens are considered correct, while the probability distributions are sometimes incorrect.
We show an example in Appendix~\ref{app:example}.

In our model, the encoder input of the pivot-target model in the source-pivot-target model is weighted embedding according to the probability distribution. 
If the probability distribution is incorrect, the pivot $\to$ target translation will also be affected.

We propose various heuristics to correct the probability distribution at inconsistent positions as follows. 
\textbf{eq-1}: set the probability of the generated token at the inconsistent position to 1.0; \textbf{add-1/0.5}: add 1.0/0.5 to the probability of the generated token;
\textbf{exc}: exchange the probability of the generated token with the maximum probability in the distribution. 
All these heuristics are designed to ensure that the generated token has the biggest probability.
In this way, the inconsistency between output tokens and the probability distribution can be solved.

In addition, beam search allows us to generate $n$ candidate pivot sentences and then $m$ target sentences for each pivot sentence. There are $n * m$  candidate target sentences for each source sentence.

\section{Experiments}
\subsection{Settings}
Following~\citet{zhang2022triangular}, we conduct extensive experiments on Chinese (Zh) - German (De) and French (Fr) - German (De) translation, with English (En) as the pivot language. 
All source-pivot and pivot-target models use Transformer base as the translation model~\citep{vaswani2017attention}.
We use SacreBLEU\footnote{SacreBLEU signature: BLEU+nrefs.1+case.mixed+\\tok.13a+smooth.exp+version.2.0.0.}~\citep{post-2018-call} as the evaluation metric. 
More details about data and hyper-parameters can be found in Appendices~\ref{app:data} and~\ref{app:parameters}.

\subsection{Baselines}
\paragraph{Direct} Train a Transformer base model directly on (source, target).

\paragraph{Pivot}  Train the source-pivot and pivot-target models independently on (source, pivot) and (pivot, target).
\paragraph{Joint Training} \citet{cheng2019joint} connect source-pivot and pivot-target models by a connection term.
They use the source-pivot model to generate the pivot translation on-the-fly and then input it into the pivot-target model to calculate the connection term loss.  
Our implementation differs from the original in that we pre-train with (source, pivot) and (pivot, target) and then fine-tune with (source, target), while they train together. Besides, our pivot translations are generated offline.

\paragraph{Step-wise Pre-training} A simple method proposed by~\citet{kim2019pivot}. First train a source-pivot model and use the source-pivot encoder to initialize the pivot-target encoder. Then train the pivot-target model with encoder frozen, and use the pivot-target model to initialize the source-target model. Finally, train the source-target model on (source, target).

\paragraph{Triangular Transfer} Triangular Transfer is a transfer-learning-based approach proposed by~\citet{zhang2022triangular}. They exploit all types of auxiliary data and design parameter freezing mechanisms to transfer the model to the source-target model smoothly.

\begin{table}[t]
\centering
\begin{tabular}{c|cc}
\toprule
Models                 & Zh-De & Fr-De \\ \midrule
Direct                 & 12.21 & 13.54 \\
Pivot                  & 13.57 & 19.05 \\
Joint Training         & 16.73 & 19.18 \\
Step-wise Pre-training* & -     & 18.49 \\
Triangular Transfer*    & 16.03 & 19.91 \\ \midrule
Ours                  & 17.02 & 19.53 \\ \bottomrule
\end{tabular}
\caption{Comparison with baselines on the test set. * represents the implementation of~\citep{zhang2022triangular}. The other models are implemented by ourselves.}
\label{tab:overall}
\end{table}

\begin{table}[t]
\centering
\begin{tabular}{c|cc}
\toprule
P.C. & Zh-De & Fr-De \\ \midrule
-           & 16.29 & 18.66 \\
eq-1                   & 16.47 & 18.84 \\
add-1                  & 16.51 & 18.87 \\
add-0.5                & 16.55 & 18.85 \\
exc                    & 16.41 & 18.74 \\ \bottomrule
\end{tabular}
\caption{The performance of different probability correction methods on the validation set. P.C. denotes probability correction.}
\label{tab:probability-correction}
\end{table}
\subsection{Overall Results}

Table~\ref{tab:overall} shows the performance of our method and baselines on Zh-De and Fr-De. 
Direct has poor performance because it only uses a small number of parallel data. 
Pivot has a significant improvement on Fr-De, but the improvement on Zh-De is limited. 
On Fr-De, our method gains 0.48 BLEU improvement over Pivot, but our improvement is not greater than Triangular Transfer. We conjecture this is due to a large number of monolingual data Triangular Transfer uses. 
On Zh-De, our method outperforms all baselines with 17.02 BLEU and outperforms Pivot by 3.45 BLEU. Our approach outperforms Joint Training, which illustrates the importance of re-normalizing the probability distribution and backpropagation. The connection term in Joint Training can only train the pivot-target model because the gradient cannot be backpropagated to the source-pivot model.

\begin{table}[t]
\centering
\begin{tabular}{c|cc|cc}
\toprule
Models & Beam & $n$-Pivot & Zh-De & Fr-De \\ \midrule
\multirow{5}{*}{Pivot} & 1    & 1       & 11.46 & 18.05 \\
       & 2    & 1       & 12.20 & 18.37 \\
       & 5    & 1       & 12.68 & 18.56 \\
       & 5    & 2       & 12.67 & 18.75 \\
       & 5    & 5       & 12.72 & 18.72 \\ \midrule
\multirow{5}{*}{Ours}       & 1    & 1       & 16.38 & 18.54 \\
       & 2    & 1       & 16.44 & 18.64  \\
       & 5    & 1       & 16.29 & 18.66 \\
       & 5    & 2       & 16.47 & 18.67 \\
       & 5    & 5       & 16.49 & 18.58 \\ \bottomrule
\end{tabular}
\caption{The performance of applying beam search for pivot language on the validation set. Beam denotes the pivot beam size and $n$-Pivot denotes the number of pivot candidates.}
\label{tab:pivot-bs}
\end{table}

\begin{table}[t]
\centering
\begin{tabular}{c|c|cc}
\toprule
$n$-Pivot & P.C. & Zh-De & Fr-De \\ \midrule
\multirow{2}{*}{2}       & -                      & 16.47 & 18.67 \\
        & eq-1                  & 16.68 & 18.91 \\ \midrule
\multirow{2}{*}{5}       & -                      & 16.49 & 18.58 \\
        & eq-1                  & 16.66 & 18.93 \\ \bottomrule
\end{tabular}
\caption{The performance of $n$-Pivot with probability correction on the validation set.}
\label{tab:probability-correction-with-npivot}
\end{table}

\subsection{Analysis}
\paragraph{Probability Correction}
As mentioned above, we propose various heuristics to correct the probability distributions of inconsistent positions. 
Table~\ref{tab:probability-correction} shows the results of different methods with a pivot beam size of 5. As we expected, all the correction heuristics improve the performance to varying degrees. Appendix~\ref{app:example} also shows the effect of probability correction on an example.

\paragraph{$n$-Pivot}
As shown in Table~\ref{tab:pivot-bs}, whether our method or Pivot, using beam search for the pivot translations can improve the final target result.
Our method gains significant improvement when using multiple pivot candidates as intermediate translation. However, it is only useful on Fr-De for Pivot. Because for Pivot, it only increases the number of candidate sentences, while for our model, it allows the model to select the less problematic one from multiple inconsistent candidates.
We further improve the performance by combining probability correction and $n$-Pivot as shown in Table~\ref{tab:probability-correction-with-npivot}.

\begin{table}[t]
\centering
\begin{tabular}{c|c|cc}
\toprule
Train-$\alpha$ & Decode-$\alpha$ & Zh-De & Fr-De \\ \midrule
\multirow{5}{*}{1.0}  & 0.7   & 8.60 &  4.27 \\
       & 0.9   & 16.58 & 18.70 \\
       & 1.0   & 16.61 & 18.93 \\
       & 1.5   & 16.34 & 18.91 \\
       & 2.0   & 16.19 & 18.90 \\ \midrule
\multirow{5}{*}{2.0}   & 0.7   & 3.56 & 1.71 \\
       & 0.9   & 15.16 & 18.54  \\
       & 1.0   & 15.49 & 18.64 \\
       & 1.5   & 15.78 & 18.83 \\
       & 2.0   & 15.85 & 18.78 \\ \bottomrule
\end{tabular}
\caption{The performance of various $\alpha$ values on the validation set.}
\label{tab:alpha}
\end{table}

\paragraph{Effect of $\alpha$}
In Table~\ref{tab:alpha}, we explore the impact of different $\alpha$ values. The best option is $\alpha=1$ for both training and decoding.
We believe this is because too sharp a distribution is not conducive to training.

\begin{table}[t]
\centering
\begin{tabular}{ccc}
\toprule
Loss   & Zh-De & Fr-De \\ \midrule
$\mathcal{L}_{target}$    & 14.50 & 18.15 \\
$\mathcal{L}$    & 16.66 & 18.93 \\ \bottomrule
\end{tabular}
\caption{Validation performance comparison to the model trained without pivot loss.}
\label{tab:beta}
\end{table}

\paragraph{Effect of Pivot Loss}
Table~\ref{tab:beta} shows the results of training with only $\mathcal{L}_{target}$ (i.e., $\beta=0$). 
We can observe the performance suffers significantly without $\mathcal{L}_{pivot}$. 
Therefore, it is necessary to construct trilingual parallel data with pseudo-pivot to supervise the training of the source-pivot model.

\section{Conclusion}
This work proposes an end-to-end approach to train the pivot-based cascaded translation model, which uses the embedding weighted according to the probability distribution as the pivot-target input rather than the embedding of a specific token. 
We also study decoding algorithms for this class of cascaded models and propose various heuristics to mitigate the inconsistency between the generated pivot tokens and probability distributions. We obtain better or comparable performance compared to previous work.

\bibliography{anthology,custom}
\bibliographystyle{acl_natbib}
\clearpage

\appendix
\begin{table*}[t!]
\centering
\begin{tabular}{ccccc}
\toprule
pair                   & source                    & train                             & valid                         & test                          \\ \midrule
Fr-En                  & WMT 2015                  & Europarl v7, News Commentary v10, & newstest2011                  & newstest2012                  \\
\multirow{2}{*}{Fr-De} & \multirow{2}{*}{WMT 2019} & News Commentary v14,              & \multirow{2}{*}{newstest2011} & \multirow{2}{*}{newstest2012} \\
                      &                           & newstest2008-2010                 &                               &                               \\
\multirow{2}{*}{En-De} & \multirow{2}{*}{WMT 2019} & Europarl v9, News Commentary v14, & \multirow{2}{*}{newstest2011} & \multirow{2}{*}{newstest2012} \\
                      &                           & Document-split Rapid corpus       &                               &                               \\
Zh-En                  & ParaCrawl                 & ParaCrawl v9                      & newsdev2017                   & newstest2017                  \\
Zh-De                  & WMT 2021                  & News Commentary v16 - dev - test  & 3k split                      & 3k split                      \\ \bottomrule
\end{tabular}
\caption{Parallel data source (from ~\citet{zhang2022triangular}).}
\label{tab:parallel-data}
\end{table*}

\section{Dataset}
\label{app:data}
We follow~\citet{zhang2022triangular} to gather parallel data and perform preprocessing.
As shown in Table~\ref{tab:parallel-data}, we gather parallel data from WMT and ParaCrawl~\citep{banon-etal-2020-paracrawl} and the training data statistics is shown in Table~\ref{tab:training-data-statistics}.
We use jieba\footnote{https://github.com/fxsjy/jieba} for Chinese (Zh) word segmentation, and Moses\footnote{https://github.com/moses-smt/mosesdecoder} scripts to normalize punctuation and tokenize for other languages. 
The data are deduplicated. 
Each language is segmented into subword units by byte pair encoding (BPE)~\citep{sennrich2016neural} with 32k merge operations. And the BPE codes and vocabularies are provided by~\citet{zhang2022triangular}. 
We remove the sentences longer than 128 subwords and clean the parallel sentences with length ratio 1.5.

\section{Hyper-parameters}
\label{app:parameters}
Our code is based on \texttt{fairseq}~\citep{ott-etal-2019-fairseq}.
We follow the hyper-parameters~\citep{vaswani2017attention} for pre-training the Transformer base models.
We train with batches of approximately $16k\cdot8$ (8 GPUs with 16k per GPU) tokens using  Adam~\citep{diederik2014adam} and enable mixed precision floating point arithmetic~\citep{micikevicius2018mixed}.
We set weight decay to 0.01 and label smoothing to 0.1 for regularization.
The learning rate warms up to $5\cdot10^{-4}$ in the first 5k steps, and then decays with the inverse square-root schedule. 
During the training of the cascaded model, the learning rate warms up to $8\cdot10^{-5}$ for Zh-De, and $8\cdot10^{-7}$ for Fr-De. 
The learning rate warms up for 500 steps, and then follows inverse square-root decay.
We use single precision floating point for fine-tuning. The batch size is $4k\cdot8$ tokens. 
$\alpha$ and $\beta$ are both set to 1.0. $\gamma$ is set to $4.0$ for Zh-De, and $1.0$ for Fr-De. We use beam size of 5 for decoding, for both pivot translation and target translation. $n$-Pivot is set to $2$ for Zh-De, and $5$ for Fr-De. Probability correction is \textbf{eq-1}.
We train all models for 300k steps on 8 NVIDIA TESLA V100 32G GPUs, and select the best checkpoints on the validation set as the final model. 
\begin{table}[!t]
\centering
\begin{tabular}{cc}
\toprule
pair  & num.  \\ \midrule
Fr-En & 29.5m \\
Zh-En & 11.9m \\
En-De & 3.1m  \\
Fr-De & 247k  \\ 
Zh-De & 189k  \\ \bottomrule
\end{tabular}
\caption{The number of sentences in parallel data.}
\label{tab:training-data-statistics}
\end{table}

\section{Example}
\label{app:example}
\begin{figure*}[t]
  \centering
  \includegraphics[width=0.95\linewidth]{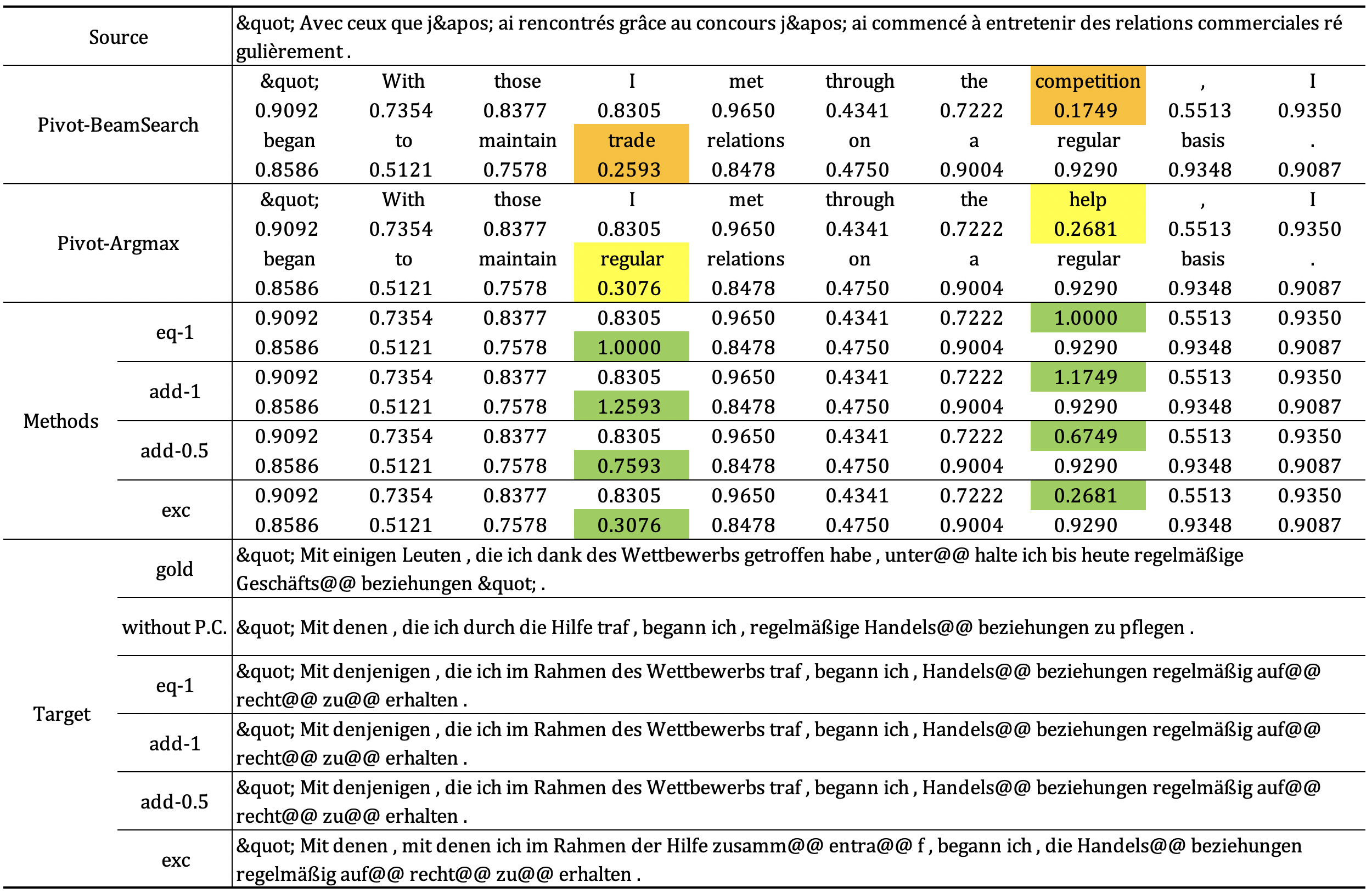}
  \caption{An example in the validation set of Fr-De.}
  \label{fig:example}
\end{figure*}

We show an example in the validation set of Fr-De in Figure~\ref{fig:example} with a pivot beam size of 5 and an $n$-Pivot of 1.
Pivot-BeamSearch is the pivot tokens and their corresponding probabilities generated by beam search.
Pivot-Argmax is the token with the highest probability at each position and the corresponding probability.
It can be seen that these tokens are different from those generated by beam search. The differences are highlighted in yellow.
For instance, \texttt{[competition]} is generated by beam search with the corresponding probability of 0.1749.
However, at this position, the most probable token is \texttt{[help]} with the corresponding probability of 0.2681.
This is what we call inconsistency. 
We need to change the incorrect probabilities of \texttt{[competition]} and \texttt{[trade]} (highlighted in orange) to the maximum. The highlighted parts in green are the corrected values.
It can be seen that the target language translation result without probability correction is affected by the incorrect probability distribution (\texttt{[help]} has the largest probability), and the translation contains \texttt{[Hilfe]}. 
However, after the probability correction, the target translation is correct, except for \textbf{exc}.
\textbf{exc} exchanges the probabilities of \texttt{[competition]} and \texttt{[help]}. The probabilities of \texttt{[competition]} and \texttt{[help]} are 0.2681 and 0.1749, respectively. The difference is not large, and \texttt{[help]} still accounts for a large proportion.

\end{document}